\documentclass[runningheads]{llncs}
\usepackage[T1]{fontenc}
\usepackage{graphicx}
\usepackage{hyperref}
\usepackage{url}
\usepackage{mathtools,amsfonts,nccmath,bbm,makecell,subcaption,multirow}
\usepackage{diagbox}
\usepackage{booktabs}

\usepackage{listings}
\usepackage{xcolor}

\DeclarePairedDelimiter{\nint}\lfloor\rceil
\begin{document}
\title{Token Sparsification for Faster \\ Medical Image Segmentation}
\author{Lei Zhou\inst{1}\thanks{Corresponding author, \email{lezzhou@cs.stonybrook.edu}} \and % index{Zhou, Lei}
Huidong Liu\inst{1,3} \and % index{Liu, Huidong}
Joseph Bae\inst{2} \and % index{Bae, Joseph}
Junjun He\inst{4} \and \\ % index{He, Junjun}
Dimitris Samaras\inst{1} \and % \index{Samaras, Dimitris}
Prateek Prasanna\inst{2}} % \index{Prasanna, Prateek}

\institute{Department of Computer Science, Stony Brook University, NY, USA \and
Department of Biomedical Informatics, Stony Brook University, NY, USA \and
Amazon, WA, USA \and Shanghai Artificial Intelligence Laboratory}

\authorrunning{Lei Zhou et al.}
\maketitle              % typeset the header of the contribution
\begin{abstract}
\textit{Can we use sparse tokens for dense prediction, e.g., segmentation?} Although token sparsification has been applied to Vision Transformers (ViT) to accelerate classification, it is still unknown how to perform segmentation from sparse tokens. To this end, we reformulate segmentation as a \textit{\underline{s}parse encoding} $\rightarrow$ \textit{token \underline{c}ompletion} $\rightarrow$ \textit{\underline{d}ense decoding} (SCD) pipeline.
We first empirically show that na\"ively applying existing approaches from classification token pruning and masked image modeling (MIM) leads to failure and inefficient training caused by inappropriate sampling algorithms and the low quality of the restored dense features.
In this paper, we propose \textit{Soft-topK Token Pruning (STP)} and \textit{Multi-layer Token Assembly (MTA)} to address these problems.
In \textit{sparse encoding}, \textit{STP} predicts token importance scores with a lightweight sub-network and samples the topK tokens. The intractable topK gradients are approximated through a continuous perturbed score distribution.
In \textit{token completion}, \textit{MTA} restores a full token sequence by assembling both sparse output tokens and pruned multi-layer intermediate ones.
% Compared to MIM which fills the pruned positions with mask tokens, \textit{MTA} produces more informative representations allowing more accurate segmentation.
The last \textit{dense decoding} stage is compatible with existing segmentation decoders, e.g., UNETR. Experiments show SCD pipelines equipped with \textit{STP} and \textit{MTA} are much faster than baselines without token pruning in both training (up to 120\% higher throughput) and inference (up to 60.6\% higher throughput) while maintaining segmentation quality. Code is available here: \url{https://github.com/cvlab-stonybrook/TokenSparse-for-MedSeg}
\keywords{Token Pruning  \and Multi-layer Token Assembly \and Medical Image Segmentation.}
\end{abstract}
\section{Introduction}
Vision Transformers (ViT)~\cite{dosovitskiy2020image} for dense prediction~\cite{zheng2021rethinking,ranftl2021vision} have achieved impressive results in tasks including medical image segmentation~\cite{hatamizadeh2022unetr}.
In general, high-resolution features~\cite{wang2020deep} preserving details are always desirable for precise segmentation. However, because of the quadratic computation complexity in self-attention~\cite{vaswani2017attention}, doubling the resolution per dimension in a 3D volume can lead to an 8$\times$ longer sequence and hence 64$\times$ more computation.
This growing computing burden can quickly surpass limited computation budgets.
Considering ViT's flexibility and great potential in masked image modeling~\cite{he2021masked,li2021benchmarking}, we %believe it is valuable to 
explore acceleration algorithms based on the standard ViT.
Recently, token sparsification ~\cite{rao2021dynamicvit,liang2021evit,meng2021adavit} has been proposed to accelerate inference in ViT for classification by dropping less important tokens. However, \textit{to the best of our knowledge, there are no ViT token sparsification approaches for segmentation}.
This leads us to ask the question: \textit{Can we use sparse tokens for dense prediction, e.g., segmentation?}

To answer the question, we reformulate segmentation as a \textit{\underline{s}parse encoding} $\rightarrow$ \textit{token \underline{c}ompletion} $\rightarrow$ \textit{\underline{d}ense decoding} (SCD) pipeline. Unlike a standard \textit{\underline{d}ense encoding} $\rightarrow$ \textit{\underline{d}ense decoding} (DD) pipeline, \textit{sparse encoding} and \textit{token completion} are required in SCD. \textit{Sparse encoding} requires learning a sparse token representation for speed and \textit{token completion} is needed to restore the full set of tokens for dense prediction.
We first examine a na\"ive realization of \textit{sparse encoding} and \textit{token completion} by applying existing approaches. Specifically, we adapt sampling methods in classification, e.g., EViT~\cite{liang2021evit} and DynamicViT~\cite{rao2021dynamicvit}, to \textit{sparse encoding}, and masked image modeling (MIM)~\cite{he2021masked,bao2021beit} to \textit{token completion}. However, we observe significantly inferior results in this SCD pipeline (See Table~\ref{tab:comb}). Next, we provide more insight into the problems of existing methods.
\\
\textbf{Problems in \textit{Sparse Encoding}}. There are two steps in this step, i.e., token score estimation and token sampling. We show that EViT's token score estimation is inappropriate for segmentation and DynamicViT's token sampling leads to training inefficiency: \textit{i)} \textbf{EViT}~\cite{liang2021evit} uses the attention weights between spatial tokens and the \texttt{[CLS]} token to estimate scores. While this is sound for classification since \texttt{[CLS]} is used for prediction, it is sub-optimal for segmentation because \texttt{[CLS]} is deprecated in the segmentation decoder. \textit{ii)} \textbf{DynamicViT}~\cite{rao2021dynamicvit} estimates token scores with a sub-network. DynamicViT frames token sampling as a series of independent binary decisions to keep or drop tokens. This does not guarantee a fixed number of sampled tokens for each training input. To fit in batch training, DynamicViT keeps all tokens in memory and masks self-attention entries, leading to training inefficiency.
\\
\textbf{Problems in \textit{Token Completion}}. Previous sparse token classification models~\cite{rao2021dynamicvit,liang2021evit} do not require \textit{token completion}. Thus, we borrow the design from MIM. MIM reconstructs full tokens from a partial token sequence by padding it to full length with learnable mask tokens and then hallucinating the masked regions from their context. While MIM is useful for pre-training, it cannot accurately restore detailed information, resulting in inferior segmentation results.

We propose \textit{Soft-topK Token Pruning (STP)} and \textit{Multi-layer Token Assembly (MTA)} to implement \textit{sparse encoding} and \textit{token completion}. \textit{i)} In \textit{sparse encoding}, \textit{STP} predicts token importance scores with a sub-network, avoiding the limitation of \texttt{[CLS]} in segmentation. STP then samples topK-scored tokens instead of making binary decisions per token separately, accelerating training by retaining only the sampled tokens in memory and computing. Motivated by subset sampling~\cite{xie2019reparameterizable,li2021differentiable,cordonnier2021differentiable}, the intractable gradients of the topK operation are approximated through a perturbed continuous score distribution.
\textit{ii)} In \textit{token completion}, the \textit{MTA} restores a full token sequence by assembling both sparse output tokens and pruned intermediate tokens from multiple layers. 
Compared to MIM that fills the pruned positions with identical mask tokens, \textit{MTA} produces more informative, position-specific representations. For \textit{dense decoding}, the SCD pipeline is compatible with existing segmentation decoders, such as UNETR.

% We expect that token pruning is most useful when segmentation targets are sparse, such as in 3D medical images.
% (See Tables~\ref{tab:btcv_sparsity}\&~\ref{tab:brats_sparsity} in the Appendix). 
We evaluate our method on two relatively sparse 3D medical image segmentation datasets, the CT Abdomen Multi-organ Segmentation (BTCV~\cite{landman2015miccai}, N=30) dataset and the MRI Brain Tumor Segmentation (MSD BraTS~\cite{antonelli2021medical}, N=484) dataset.  On both tasks, {\small STP+MTA+UNETR} matches the UNETR baseline while providing significant computing savings with large token pruning ratios. On BraTS, {\small STP+MTA+UNETR} accelerates segmentation inference/training throughput by 60.6\%/120\% and achieves the same segmentation accuracy. On BTCV, {\small STP+MTA+UNETR} increases inference/training throughput by 24.1\%/97.36\% while maintaining performance. In summary, our contributions are: 
\begin{itemize}
    \item To the best of our knowledge, we are the first to use token pruning/dropping for ViT-based medical image segmentation.
    \item Based on subset sampling, our proposed  \textit{Soft-topK Token Pruning (STP)} module can be flexibly incorporated into a standard ViT to prune tokens with greater efficiency while maintaining accuracy.
    \item We propose \textit{Multi-layer Token Assembly (MTA)} to recover a full set of tokens, i.e., a dense representation, from a sparse set. \textit{MTA}  preserves high-detail information for accurate segmentation.
    \item We show that {\small STP+MTA+UNETR} maintains performance compared with UNETR with much less computation on two 3D medical image datasets.
\end{itemize}

\section{Methodology}
\label{method}
Generally, a segmentation model consists of an encoder and a decoder.
Our goal is to accelerate the ViT segmentation encoder. To this end, we reformulate segmentation as a \textit{\underline{s}parse encoding} $\rightarrow$ \textit{token \underline{c}ompletion} $\rightarrow$ \textit{\underline{d}ense decoding} (SCD) pipeline. \textit{Sparse encoding} learns a sparse token representation for acceleration; \textit{token completion} restores the full tokens for dense prediction; \textit{dense decoding} predicts the segmentation mask from dense features. We first recap Vision Transformers and then illustrate the three components in the SCD pipeline.
\begin{figure}[t]
    \centering
    \includegraphics[width=0.94\linewidth]{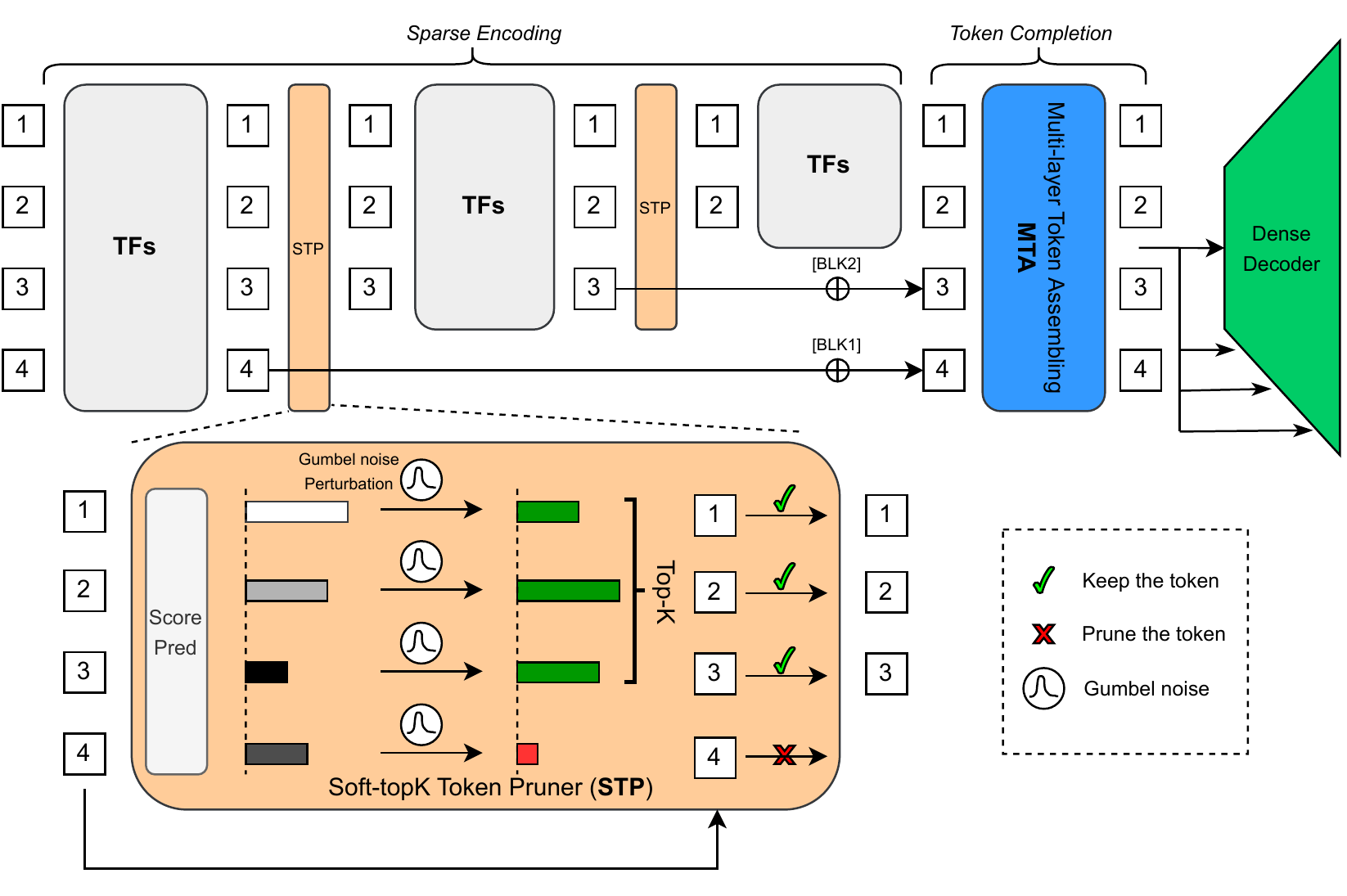}
    \caption{\textbf{Sparse Token Segmentation Pipeline}. 
    We reformulate segmentation as a \textit{sparse encoding} $\rightarrow$  \textit{token completion} $\rightarrow$ \textit{dense decoding} pipeline. In \textit{sparse encoding}, we design a \textit{Soft-topK Token Pruning (STP)} module. In the forward pass, \textit{STP} performs  topK sampling on perturbed scores. In the backward pass, \textit{STP} approximates the intractable gradient with a continuous Gumbel Softmax estimation. In \textit{token completion}, we propose \textit{Multi-layer Token Assembly (MTA)} to assemble both the output sparse tokens and the pruned intermediate ones to restore the complete tokens.
    % To identify which block the pruned tokens are from, we augment them with block-wise learnable \texttt{[BLK]} tokens. 
    In \textit{dense decoding}, we avoid the intermediate sparse tokens by taking all inputs from the output of \textit{MTA}.
    In this simplified figure, we visualize token pruning as dropping the last token. However, in practice pruned tokens are selected according to predicted scores.
    }
    \label{fig:pipeline}
\end{figure}
\\
\noindent\textbf{Preliminary: Vision Transformers.}
Vision Transformers treat an image/volume as a sequence of tokens.
In the case of 3D medical images, a 3D volume $\mathbf{x} \in \mathbb{R}^{H \times W \times D \times C_{in}}$ is first reshaped to 
a sequence of flattened patches $\mathbf{x}_p \in \mathbb{R}^{N \times (P^3 \times C_{in})}$
where $H \times W \times D$ is the spatial size, $C_{in}$ is the input channel, $P \times P \times P$ is the patch size, and $N = HWD/P^3$ is the  sequence length, i.e., the number of patches.
All the patches are then projected linearly to a $C$-dimensional token space,  with position embeddings added to the projected patches.
These patch tokens, together with a learnable prepended \texttt{[CLS]} token, are denoted as $\mathbf{z}_0 \in \mathbb{R}^{(1+N) \times C}$.
$\mathbf{z}_0$ are further processed by $L$ Transformer blocks sequentially. Each block consists of a multi-head self-attention (MSA) module and an MLP. We denote the tokens output from the ${i}$th Transformer block as $\mathbf{z}_i \in \mathbb{R}^{(1+N)\times C}$. For the segmentation task, before feeding the output $\mathbf{z}_L$ of the encoder to the decoder, we drop the \texttt{[CLS]} token and project the non\texttt{-[CLS]} token sequence $\mathbf{z}_L^{[1:N]} \in \mathbb{R}^{N\times C}$ back to the original 3D feature map $\mathbf{x}_L \in \mathbb{R}^{H/P \times W/P \times D/P \times C}$.

\subsection{Sparse Encoding: Soft-topK Token Pruning (STP)}
\label{txt:enc}
We build our sparse encoder on a ViT without modifying the self-attention module. Instead, we propose a learnable plug-and-play \textit{Soft-topK Token Pruning (STP)} module.
% such that the subsequent Transformer blocks have fewer tokens as input and are thus sped up.
Compared to EViT \& DynamicViT, our \textit{STP}, as shown in the lower half of Fig.~\ref{fig:pipeline}, estimates token scores more effectively and can be trained efficiently. \textit{STP} can be inserted between two Transformer blocks $\mathtt{TF}_i$ and $\mathtt{TF}_{i+1}$. Receiving as input the token sequence $\mathbf{z}_i \in \mathbb{R}^{N_i \times C}$ from $\mathtt{TF}_i$, \textit{STP} prunes tokens with a ratio $r$ and passes the remaining tokens $\mathbf{z}_i^\prime \in \mathbb{R}^{\nint{(1-r)N_i}\times C}$ to $\mathtt{TF}_{i+1}$. In particular, \textit{STP} consists of token-wise score estimation and token sampling. To be concise, we change the notation of number of tokens from $N_i$ to $n$.
\\
\textbf{Token Score Estimation.} To decide which tokens to keep or prune, we introduce a lightweight sub-network $s_\theta: \mathbb{R}^{n\times C}\rightarrow \mathbb{R}^{n}$ to predict the token importance scores $\mathbf{s}$, where $\theta$ are the network parameters. The architecture of $s_\theta$ is designed to aggregate both the local and global features, similarly to \cite{rao2021dynamicvit}. The global feature is simply obtained by average pooling over all the tokens.
\begin{equation}
    \mathbf{s} = s_\theta(\mathbf{z}) = \mathtt{Sigmoid}\bigg(\mathtt{MLP_2}\Big([\mathbf{z},\mathtt{AvgPool}\big(\mathtt{MLP_1}(\mathbf{z})\big)]\Big)\bigg)
\end{equation}
\\
\textbf{Straight-through Gumbel Soft TopK Sampling.} 
\label{st-gumbel} 
Given a token pruning ratio $r$, \textit{STP} needs to select $K = \nint{(1-r)n}$ tokens out of $n$ to keep. After predicting the scores $\mathbf{s}$, we re-interpret each score value $\mathbf{s}_i$ as the probability of the $i$-th token ranking in the topK. We formulate this process as sampling a binary policy mask $\mathbf{M} \in \{0,1\}^{n}$ from the predicted probabilities where $\mathbf{M}$ is subject to $\mathtt{sum}(\mathbf{M}) = K$. $\mathbf{M}_i = 1$ indicates keeping the $i$-th token while $\mathbf{M}_i = 0$ indicates pruning. However, such  discrete sampling is non-differentiable. To overcome the problem, we relax the sampling of discrete topK masks to a continuous approximation, the Gumbel-Softmax distribution:
% (Please refer to Paragraph Gumbel Softmax~\ref{gumbel-softmax-related} in Related Work for more details). 
% This approximation can be summarized as,
\begin{equation}
\label{eq:softopk}
\underbrace{\mathbf{M}_i = \mathbbm{1}_{\text{topK}}(\mathrm{log}(s_i)+g_i)}_\text{forward} \xleftarrow{\text{approx}} \underbrace{\Tilde{\mathbf{M}}_i = \frac{\mathrm{exp}((\mathrm{log}(s_i)+g_i)/\tau)}{\sum_{j=1}^{n}\mathrm{exp}((\mathrm{log}(s_j) + g_j)/\tau)}}_\text{backward}
\end{equation}
where $\mathbbm{1}_{\text{topK}}$ is an indicator function of whether the input perturbed score is among the topK of all $n$ perturbed scores, $\{g\}_n$ are i.i.d samples from the $\textrm{Gumbel}(0, 1)$ distribution\footnote{$\mathrm{Gumbel}(0,1)$ samples are drawn by sampling $-\mathrm{log}(-\mathrm{log}\;u)$ where $u \sim \mathrm{Uniform}(0, 1)$}. 
While training, we forward \textit{STP} to sample the topK tokens based on the discrete $\mathbf{M}$ but backward with the gradient approximated by the continuous $\Tilde{\mathbf{M}}$. We call this Straight-through (ST) Gumbel Soft TopK Sampling. During inference, we perform normal topK selection based on predicted scores without Gumbel noise perturbation for deterministic inference.

\subsection{Token Completion: Multi-layer Token Assembly (MTA)}
\label{txt:compl}
The output of the STP-ViT encoder is sparse. Thus, before passing the output to the decoder, we need to first restore the complete tokens. A straightforward solution can be obtained from Masked Image Modeling (MIM)~\cite{bao2021beit,he2021masked}. MIM reconstructs an image from random partial image patches. It first pads the sparse token set with learnable \texttt{[MASK]} tokens up to its full length. Then the padded tokens are forwarded through Transformer blocks to reconstruct the masked regions. However, MIM is mostly utilized for pre-training which focuses more on semantic hallucination rather than accurate detail restoration. Thus, it is sub-optimal for segmentation tasks that require assigning labels to pixels accurately.

We propose \textit{Multi-layer Token Assembly (MTA)} to restore dense features by assembling both the outputted sparse tokens and the pruned intermediate tokens from multiple layers. Suppose we insert three \textit{STP}s, $\{\mathit{STP}_1, \mathit{STP}_2, \mathit{STP}_3\}$, after different Transformer blocks in a ViT. We denote the token sets pruned by the three \textit{STP}s as $\{\bar{\mathbf{z}}_1, \bar{\mathbf{z}}_2, \bar{\mathbf{z}}_3\}$. We concatenate these pruned tokens with the final output $\mathbf{z}_L$ and rearrange them to their original spatial order. Then, we add three learnable block tokens $\{\mathtt{[BLK_1], [BLK_2], [BLK_3]}\}$ to the corresponding pruned tokens to indicate which block each token is pruned from. Finally, we introduce sin-cos position embeddings $\mathbf{E}_{pos}$ to all the tokens and forward them through Transformer blocks. The completion process can be summarized as follows:
\begin{equation}
\mathbf{z}_\mathrm{compl} = \mathtt{TF}(\mathtt{rearrange}(\big[\bar{\mathbf{z}}_1 + \mathtt{[BLK_1]}, \bar{\mathbf{z}}_2 + \mathtt{[BLK_2]}, \bar{\mathbf{z}}_3 + \mathtt{[BLK_3]}, \mathbf{z}_L\big]) + \mathbf{E}_{pos})
\end{equation}
\subsection{Dense Decoding \& Optimization}
\label{txt:dec}
As our goal is to design an acceleration method that is agnostic to decoder designs, designing a new segmentation decoder is beyond the scope of this paper. Thus, we couple the SCD pipeline with existing segmentation decoders.
However, certain segmentation decoders, e.g., UNETR, require inputs from multiple layer outputs from the encoder, which causes problems because intermediate features are still sparse. Motivated by  recent research on the non-hierarchical feature pyramid~\cite{li2022exploring}, we use the output $\mathbf{z}_\mathrm{compl}$ of the completion network to replace all the intermediate features required by the segmentation head, as shown in Fig.~\ref{fig:pipeline}.
% In our experiments, this approach has  performed well.

Unlike DynamicViT, we do not introduce additional loss functions for token pruning. We optimize all segmentation models  by segmentation loss. We adopt a combination of cross entropy and Dice loss. Both loss weights are set to 1.

\section{Experiments}
\label{exp}
\subsection{Dataset Description}
\label{exp:setup}
We evaluate on two benchmark 3D medical segmentation datasets with sparse targets.
% (See Tables~\ref{tab:btcv_sparsity} \&~\ref{tab:brats_sparsity} in Appendix). 
The tasks are CT multi-organ and MRI Brain tumor segmentation.\\
\textbf{CT Multi-organ Segmentation (BTCV).} The BTCV~\cite{landman2015miccai} (Multi Atlas Labeling Beyond The Cranial Vault) dataset consists of 30 subjects with abdominal CT scans where 13 organs were annotated under the supervision of board-certified radiologists. Each CT volume has $85\sim 198$ slices of $512 \times 512$ pixels, with a voxel spatial resolution of ($0.54 \times 0.98 \times [2.5 \sim 5.0]$ $mm^3$). For comparison convenience, we follow~\cite{chen2021transunet,transunet-github} to split the 30 cases into 18 for training and 12 for validation. Hyper-parameters are selected via 3-fold cross validation in the training set. We report the average DSC (Dice Similarity Coefficient) and 95\% Hausdorff Distance (HD95) on 8 abdominal organs (aorta, gallbladder, spleen, left kidney, right kidney, liver, pancreas, spleen, stomach) to align with~\cite{chen2021transunet}.\\
\textbf{MRI Brain Tumor Segmentation (BraTS).}
The Medical Segmentation Decathlon (MSD)~\cite{antonelli2021medical} BraTS dataset has 484 multi-modal (FLAIR, T1w, T1-Gd and T2w) MRI scans. The ground-truth segmentation labels include peritumoral edema, GD-enhancing tumor and the necrotic/non-enhancing tumor core. The performance is measured on three recombined regions, i.e., tumor core, whole tumor and enhancing tumor. We randomly split the dataset into training (80\%), validation (15\%), and test (5\%) sets. We report average DSC and HD95.
\subsection{Implementation Details}
\label{sec:imp}
Our method is implemented in PyTorch~\cite{paszke2019pytorch} and MONAI~\cite{MONAI_Consortium_MONAI_Medical_Open_2020} on a single NVIDIA A100.
Our encoder is based on a ViT-Base model. Three \textit{STP} modules are inserted after the 3rd, 6th, and 9th Transformer blocks in ViT-B. We follow UNETR~\cite{hatamizadeh2022unetr} on data processing. 
For BTCV, we clip the raw values between -958 and 326, and re-scale the range between -1 and 1. 
% During training, we randomly flip and crop a $96\times 96\times 96$ 3D volume as the input. 
For BraTS, we perform an instance-wise normalization over the non-zero region per channel. 
% During training, we randomly flip and crop a $128\times 128\times128$ 3D volume as the input.
% Please refer to Sec.~\ref{apx:imple} for data processing in the Appendix. 
For training, we set the batch size to 2 and the initial learning rate to 1.3e-4. We use AdamW as the optimizer and adopt layer-wise learning rate decay (ratio=0.75) to improve training. For inference, we use a sliding window with an overlap of 50\%.
% Experiments were run on a single NVIDIA A100.

\subsection{Results}
\label{subsec:topunetr}
\textbf{Na\"ive Combination of EViT/DynamicViT+MIM.}
We first test the straightforward approach of applying EViT/DynamicViT to \textit{sparse encoding} and MIM to \textit{token completion}. We use UNETR as the segmentation decoder. In Table~\ref{tab:comb}, EViT/DynamicViT + MIM fails to perform dense prediction for a very high pruning ratio $r = 0.9$ on BTCV. This justifies our efforts in this paper to accelerate sparse token segmentation models while maintaining performance.

\begin{table}[t]
    \centering
    \scalebox{0.8}{
    \begin{tabular}{@{\extracolsep{4pt}}cc|ccc}
        \toprule
        \multicolumn{2}{c}{\multirow{2}{*}{\makecell{DSC(\%) on BTCV\\(pruning ratio $r = 0.9$)}}} &  \multicolumn{3}{c}{\textit{sparse encoding}} \\
        \cmidrule(lr){3-5}
        & & DynamicViT~\cite{rao2021dynamicvit} & EViT~\cite{liang2021evit} & STP (ours) \\
        \midrule
        \multirow{2}{*}{\makecell{\textit{token completion}}} & MIM~\cite{bao2021beit,he2021masked} & 24.35 (single run) & 18.64 (single run) & 44.71 (single run) \\
        & MTA (ours) & $80.24\pm 0.34$ & $78.62 \pm 0.10$ & $82.18\pm 0.12$ \\
        \bottomrule
    \end{tabular}
    }
    \caption{\textbf{Performance of existing approaches on BTCV}. We first examine the performance of the na\"ive combination of existing approaches. 
    % EViT/DynamicViT for token sampling, and mask image modeling (MIM) for token completion. 
    For a large pruning ratio $r=0.9$ on BTCV, MIM fails to perform segmentation effectively. Even with our proposed MTA instead of MIM, EViT and DynamicViT still perform worse than our STP. We report the \texttt{mean} and \texttt{std} on three random runs unless otherwise stated. Please see Sec.~\ref{exp} for more analysis.}
    \label{tab:comb}
\end{table}
\noindent \textbf{Our Approach: STP+MTA.} We evaluate the efficiency of our \textit{Soft-topK Token Pruning (STP)} and \textit{Multi-layer Token Assembly (MTA)} on the BTCV and BraTS datasets based on UNETR. We measure the efficiency by profiling the throughput(image/s) and MAC number (Multiply–accumulate operations) for each model variant. The throughput is measured on a NVIDIA A100 GPU with batch size 1. MACs are computed by measuring the forward complexity of a single image.
We present the results in Table~\ref{tab:topunetr}. On BraTS, with an input size of $(128\times 128\times 128)$, our {\small STP+MTA+UNETR} ($r=0.75$) maintains performance while significantly increasing inference throughput by 60.8\%. On BTCV, with an input size of $(96 \times 96\times 96)$, {\small STP+MTA+UNETR} ($r=0.9$) can maintain performance while the corresponding inference throughput increases by 24.1\%.
Our method also increases training efficiency. The training throughput on BTCV increases from 2.65 imgs/s to 5.23 imgs/s by 97.36\%. The training throughput on BraTS increases from 0.75 imgs/s to 1.65 imgs/s by 120\%.

\begin{table}[t]
    \centering
    \scalebox{0.8}{
    \begin{tabular}{@{\extracolsep{4pt}}ccccccc}
        \hline
        \multirow{2}{*}{Method} & \multicolumn{2}{c}{MSD BraTS} & \multirow{2}{*}{\makecell{Encoder \\ Throughput(img/s)}} & \multirow{2}{*}{\makecell{Throughput \\ (img/s)}} & \multirow{2}{*}{MACs(G)} \\
        \cline{2-3}
        & DSC$\uparrow$ & HD95$\downarrow$ \\
        \hline
        UNETR & 75.44 & 8.89 & 7.10 & 4.85 & 824.38\\
        {\small STP+MTA+UNETR} & \textbf{75.79} & \textbf{8.31} & 20.04 & 7.79 \textcolor{blue}{{\footnotesize $(+60.6\%)$}} & 428.28  \\
        \hline
    \end{tabular}}
    \scalebox{0.8}{
    \begin{tabular}{@{\extracolsep{4pt}}cccccc}
        \hline
        \multirow{2}{*}{Method} &  \multicolumn{2}{c}{BTCV} & \multirow{2}{*}{\makecell{Encoder \\ Throughput(img/s)}} & \multirow{2}{*}{\makecell{Throughput\\(img/s)}} & \multirow{2}{*}{MACs(G)} \\
        \cline{2-3}
        & DSC$\uparrow$ & HD95$\downarrow$ \\
        \hline
        UNETR & $80.78\pm 0.34$ & $\textbf{15.90}\pm 1.01$ & 30.30 & 16.18 & 273.45 \\
        {\small STP+MTA+UNETR} & $\textbf{82.18}\pm 0.12$ & $19.85\pm 1.12$ & 57.31 & 20.08 \textcolor{blue}{{\footnotesize $(+24.1\%)$}} & 146.63 \\
        \hline
    \end{tabular}}
    \caption{\textbf{{\small STP+MTA+UNETR} vs. UNETR performance comparison}. Based on the same ViT scale and patch size, our proposed {\small STP+MTA+UNETR} can maintain performance while significantly reducing computation by a large margin. We report the \texttt{mean} and \texttt{std} of three random runs on BTCV. Please refer to Sec~\ref{subsec:topunetr} for more details on the experimental setting and analysis.}
    \label{tab:topunetr}
\end{table}
\noindent \textbf{Sparse Encoding: STP vs. EViT/DynamicViT.}
EViT~\cite{liang2021evit} and DynamicViT~\cite{rao2021dynamicvit} were initially designed for classification. 
% Therefore, there are no official implementations for segmentation. 
Thus, we need to adapt EViT/DynamicViT for comparison. To constrain the pruning ratio in DynamicViT, we add the ratio loss function $\mathcal{L}_{ratio}$ with a weight of $\mathcal{\lambda}_{ratio} = 2$ following~\cite{rao2021dynamicvit}. In EViT, we take the \texttt{[CLS]} attention weights from the Transformer block as the token scores and use topK for sampling. As shown in Table~\ref{tab:tpsota}, our STP-ViT performs the best. The inferiority of DynamicViT could be caused by \textit{i}) mismatch between the training (variable number of pruned tokens) and testing phases (fixed number of pruned tokens) and \textit{ii}) more hyper-parameters (e.g., $\mathcal{\lambda}_{ratio}$). The performance drop in EViT indicates that the \texttt{[CLS]} attention scores are not suitable for representing the true token importance in segmentation.
% \textcolor{red}{Please refer to Sec.~{apx:devit} in Appendix for more implementation details.}
\\
\textbf{Token Completion: MTA vs. MIM.} 
% Token completion is a unique component in a SCD pipeline. 
We implement a baseline inspired by MIM~\cite{bao2021beit,he2021masked}. As Table~\ref{tab:resuse} shows, MIM-style completion fails ($44.71\%$) with a high pruning ratio $r=0.9$.
Our results suggest that pruned token reuse in \textit{MTA} plays an important role in a highly sparse token segmentation framework.
\\
\textbf{Token Pruning Ratio in STP.}
\label{par:pr}
We ablate the pruning ratio in Table~\ref{tab:pr}.
\textit{STP} is robust to a wide range of pruning ratios $[0.25, 0.9]$. Thus, our {\small STP+MTA+UNETR} can adopt a high pruning ratio to reduce computation by a large margin. Although our method achieves higher DSC on BTCV than UNETR, the HD95 is worse. We speculate that HD95 is more sensitive to the boundary segmentation results and that token pruning may lead to sub-optimal boundary prediction.

\begin{table}[t]
    \centering
    \scalebox{0.8}{
    \begin{tabular}{@{\extracolsep{4pt}}cccccccc}
        \hline
        \multirow{2}{*}{\makecell{Pruning Ratio \\ $r$}} &  \multicolumn{2}{c}{BTCV} & \multicolumn{2}{c}{BraTS} & \multirow{2}{*}{\makecell{Encoder \\ Throughput}} & \multirow{2}{*}{Throughput} & \multirow{2}{*}{MACs(G)} \\
        \cline{2-3} \cline{4-5}
        & DSC$\uparrow$ & HD95$\downarrow$ & DSC$\uparrow$ & HD95$\downarrow$ \\
        \hline
        baseline & $80.78\pm 0.34$ & $15.90\pm 1.01$ & 75.44 & 8.89 & 7.10 & 4.85 & 824.38 \\
        \hline
        0.25 & $81.56\pm 0.16$ & $19.65\pm 3.25$ & 75.50 & 7.98 & 11.77 & 6.12 & 631.75 \\
        0.50 & $81.81 \pm 0.59$ & $\mathbf{15.78}\pm 1.01$ & 75.02 & \textbf{7.40} & 17.34 & 7.35 & 497.97 \\
        0.75 & $81.95 \pm 0.18$ & $16.37 \pm 5.41$ & \textbf{75.79} & 8.31 & 20.04 & 7.79 & 428.28 \\
        0.9 & $\mathbf{82.18} \pm 0.12$ & $19.85 \pm 1.12$ & 75.32 & 8.04 & 21.63 & 8.04 & 404.14 \\
        \hline
    \end{tabular}}
\caption{\textbf{Ablation on the Pruning Ratio $r$.} STP shows robustness to a wide range of pruning ratios ($0.25\rightarrow 0.9$) in terms of DSC. Different datasets have different optimal pruning ratios.
Refer to Sec~\ref{par:pr} for more details. We report the \texttt{mean} and \texttt{std} of three random runs on BTCV unless otherwise stated.}
\label{tab:pr}
\end{table}

\begin{table}[t]
\begin{subtable}{0.48\linewidth}
\centering
{
\scalebox{0.8}{
\begin{tabular}{cc}
\makecell{Encoder} & DSC \\
\hline
DynamicViT & $80.24 \pm 0.34$ \\
EViT & $78.62 \pm 0.10$ \\
STP-ViT (Ours) & $\mathbf{82.18} \pm 0.12$
\end{tabular}}}
\caption{Comparison with DynamicViT\&EViT}\label{tab:tpsota}
\end{subtable}
\begin{subtable}{0.48\textwidth}
\centering
{
\scalebox{0.8}{
\begin{tabular}{cc}
\makecell{Token Completion} & DSC \\
\hline
MIM & 44.71 (single run) \\
% MIM & 81.95 \\
MTA (ours) & $\mathbf{82.18} \pm 0.12$ \\
\end{tabular}}}
\caption{Token Completion Methods}\label{tab:resuse}
\end{subtable}
\begin{subtable}{0.48\textwidth}
\centering
{
\scalebox{0.8}{
\begin{tabular}{cc}
\makecell{Perturbation} & DSC \\
\hline
No (ST TopK) & $81.67 \pm 0.21$ \\
Yes (ours) & $\mathbf{82.18}\pm 0.12 $ \\
\end{tabular}}}
\caption{Gumbel Perturbation}\label{tab:pert}
\end{subtable}
\begin{subtable}{0.48\textwidth}
\centering
{
\scalebox{0.8}{
\begin{tabular}{lc}
$\tau$ & DSC \\
\hline
0.01 & $81.36 \pm 0.15$ \\
0.1 & $82.06 \pm 0.22$ \\
1 (ours) & $\mathbf{82.18} \pm 0.12$ \\
\end{tabular}}}
\caption{Temperature $\tau$ in \textit{STP}}\label{tab:tau}
\end{subtable}
\caption{\textbf{Ablation studies on BTCV}. In (a), we compare \textit{STP} with DynamicViT and EViT. \textit{STP} achieves better performance. In (b), we compare our proposed \textit{MTA} with MIM where MIM performs much worse than MTA. In (c), we demonstrate that Gumbel perturbation is beneficial. 
In (d), we ablate different $\tau$ values. $\tau=0.1$ and $\tau=1$ perform similarly while $\tau=0.01$ performs worse. We report the \texttt{mean} and \texttt{std} of three random runs unless otherwise stated.}\label{tab:1}
\end{table}
\noindent \textbf{Temperature $\tau$ in STP.}
We ablate temperature $\tau$ in Eq.~\ref{eq:softopk} in Table~\ref{tab:tau}. According to ~\cite{jang2016categorical}, a small temperature leads to a large variance of gradients and vice versa. We tried three different $\tau$ values $\{0.01, 0.1, 1\}$. Experiments show $\tau=0.1$ and $\tau=1$ perform similarly while $\tau=0.01$ performs worse.
\\
\textbf{Noise Perturbation in STP.}
In \textit{Soft-topK Token Pruning (STP)}, we design a straight-through (ST) Gumbel soft topK algorithm for sampling. \textit{STP} forward process can be split into three steps, i.e., score prediction, Gumbel perturbation, and topK sampling. In Table~\ref{tab:pert}, we ablate the Gumbel perturbation on BTCV by evaluating a straight-through (ST) topK variant. Note that we do not add Gumbel noise during inference, to ensure that the model performs deterministically for inference. For the ST topK variant, we also remove the Gumbel noise perturbation from the training phase. With a pruning ratio $r=0.9$, results show that the Gumbel perturbation is beneficial. It is worth noting that the ST topK variant without perturbation  also achieves a competitive result.

\begin{table}[t!]
    \centering
    \scalebox{0.63}{
    \begin{tabular}{cccccccccc}
    \hline
    Framework & DSC$\uparrow$/HD95$\downarrow$ & Aorta & Gallbladder & Kidney(L) & Kidney(R) & Liver & Pancreas & Spleen & Stomach \\
    \hline
    V-Net~\cite{milletari2016v} & 68.81/- & 75.34 & 51.87 & 77.10 & 80.75 & 87.84 & 40.05 & 80.56 & 56.98 \\
    DARR~\cite{fu2020domain} & 69.77/- & 74.74 & 53.77 & 72.31 & 73.24 & 94.08 & 54.18 & 89.90 & 45.96 \\
    U-Net(R50)~\cite{ronneberger2015u} & 74.68/36.87 & 84.18 & 62.84 & 79.19 & 71.29 & 93.35 & 48.23 & 84.41 & 73.92 \\
    AttnUNet(R50)~\cite{schlemper2019attention} & 75.57/36.97 & 55.92 & 63.91 & 79.20 & 72.71 & 93.56 & 49.37 & 87.19 & 74.95 \\
    TransUNet~\cite{chen2021transunet} & 77.48/31.69 & 87.23 & 63.13 & 81.87 & 77.02 & 94.08 & 55.86 & 85.08 & 75.62 \\
    \hline
    UNETR (PatchSize=16) & 78.83/25.59 & 85.46 & 70.88 & 83.03 & 82.02 & 95.83 & 50.99 & 88.26 & 72.74 \\
    UNETR (PatchSize=8) & 80.78/\textbf{15.90} & 88.59 & 70.97 & 83.38 & 83.76 & 95.52 & 59.76 & 88.53 & 74.30 \\
    {\small STP+MTA+UNETR} (PatchSize=8) & \textbf{82.18}/19.85 & 89.23 & 73.60 & 85.66 & 83.65 & 95.59 & 62.17 & 88.84 & 77.37 \\
    \hline
    \end{tabular}}
    \caption{Comparison with other methods on BTCV.}
    \label{tab:sota}
\end{table}
\begin{figure}[t!]
    \centering
    \includegraphics[width=0.9\textwidth]{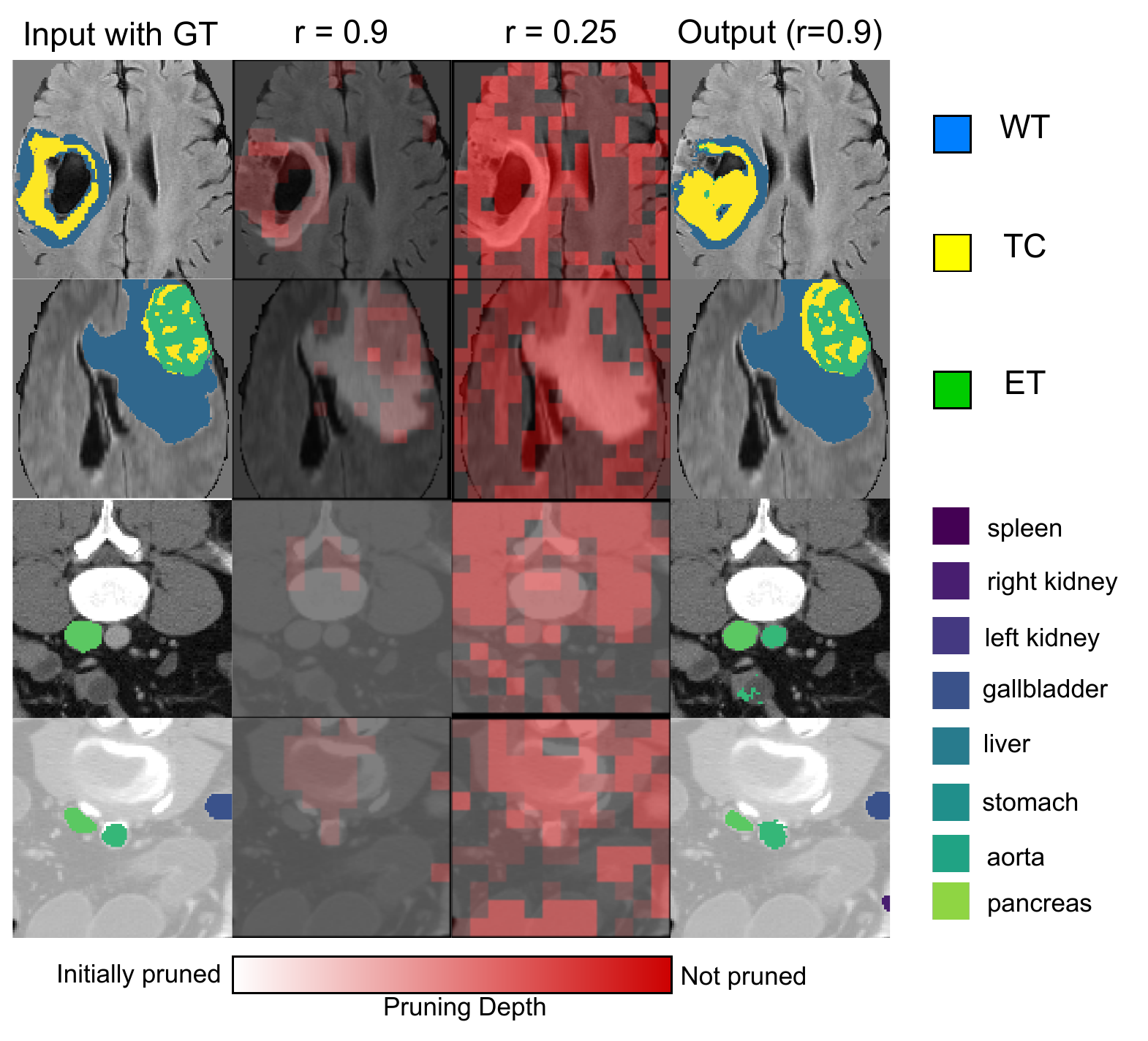}
    \caption{Ground truth and model outputs on BraTS (first two rows) and BTCV (last two rows). We visualize the depth at which tokens are pruned under high (r=0.9) and low (r=0.25) pruning ratios (red shading in columns 2 and 3). Tokens that are immediately dropped are not shaded, whereas darker red shading indicates the pruning of tokens in later layers.}
    \label{fig:policy}
\end{figure}
\noindent \textbf{Pruning Policy Visualization.}
We visualize the pruning policy for both brain tumors and abdominal organs in Fig.~\ref{fig:policy} under two extreme pruning ratios, the highest one at $r=0.9$ and the lowest at $r=0.25$. We use shades of red to denote the depth at which tokens are pruned. Patches (tokens in ViT) with no red overlap are pruned by the very first \textit{STP}, whereas patches with the deepest red color are kept in ViT until the last. In Fig~\ref{fig:policy}, with $r=0.9$, most tokens are dropped at a very early stage. Some tokens around the brain tumor, especially at tumor boundaries, are never pruned. When the ratio decreases to $r=0.25$, more patches are kept and still cluster around the target tumor region.
\\
\textbf{Class-wise Comparison with Others on BTCV.} We show class-wise results of UNETR, {\small STP+MTA+UNETR}, and other methods in Table~\ref{tab:sota}. 

{\small STP+MTA+UNETR} shows improvement over a series of methods on BTCV. Note that current SOTA methods~\cite{zhou2021nnformer,wu2022d,tang2022self} rely on either stronger priors (window attention) or SSL pre-training. However, our goal is accelerating standard ViT-based segmentation instead of purely pursuing increased performance.

\section{Conclusion and Future Work}
\label{sec:con}
We introduced a ViT-based sparse token segmentation framework for medical images. First, we proposed a \textit{Soft-topK Token Pruning} (STP) module to prune tokens in ViT. 
STP can speed up ViTs in both training and inference phases. To produce a full set of tokens for dense prediction, we proposed \textit{Multi-layer Token Assembly (MTA)} that recovers a complete set of tokens by assembling both output  and intermediate tokens from multiple layers. In our  3D medical image experiments {\small STP+MTA+UNETR} speeds up the UNETR baseline significantly while maintaining segmentation performance.
Accelerating the decoder, which also plays a big role in the inference speed, is left for future work.

\section*{Acknowledgement}
The reported research was partly supported by NIH award $\#$ 1R21CA258493-01A1, NSF awards IIS-2212046 and IIS-2123920, and Stony Brook OVPR seed grants. The content is solely the responsibility of the authors and does not necessarily represent the official views of the National Institutes of Health

\bibliographystyle{splncs04}
\bibliography{mybib}

\end{document}